%% file: main.tex
\documentclass[conference]{IEEEtran}
\IEEEoverridecommandlockouts
\usepackage{cite}
\usepackage{amsmath,amssymb,amsfonts}
\usepackage{algorithmic}
\usepackage{graphicx}
\usepackage{textcomp}
\usepackage{xcolor}
\usepackage[hidelinks]{hyperref}
\def\BibTeX{{\rm B\kern-.05em{\sc i\kern-.025em b}\kern-.08em
    T\kern-.1667em\lower.7ex\hbox{E}\kern-.125emX}}
\begin{document}

\title{A Comparative Analysis of Transformer and LSTM Models for Detecting Suicidal Ideation on Reddit
}
%-------------------------------------
\author{\IEEEauthorblockN{Khalid Hasan}
\IEEEauthorblockA{\textit{Department of Computer Science} \\
\textit{Missouri State University}\\
Springfield, Missouri, USA \\
kh597s@missouristate.edu}
\and
\IEEEauthorblockN{Jamil Saquer}
\IEEEauthorblockA{\textit{Department of Computer Science} \\
\textit{Missouri State University}\\
Springfield, Missouri, USA \\
jamilsaquer@missouristate.edu}
}
%-------------------------------
% \author{\IEEEauthorblockN{Author1}
% %\IEEEauthorblockA{\textit{Department} \\
% %\textit{University}\\
% %Address \\
% %email}
% \and
% \IEEEauthorblockN{Author2}
% %\IEEEauthorblockA{\textit{Department} \\
% %\textit{University}\\
% %Address \\
% %email}
% }

\maketitle

\input{abstract}


\input{Sections/introduction}

\input{Sections/related_work}
\input{Sections/problem_definition}
\input{Sections/methodology}

\input{Sections/results}
% \input{Sections/ethics_and_limitations}
\input{Sections/conclusion}

\bibliographystyle{IEEEtran}
\bibliography{bibtex}

\end{document}

%% file: abstract.tex
\begin{abstract}
Suicide is a critical global health problem involving more than 700,000 deaths yearly, particularly among young adults. Many people express their suicidal thoughts on social media platforms such as Reddit.
%This makes social media platforms like Reddit an important domain for people to share suicidal thoughts, hence, effective mechanisms are required for detecting suicidal thoughts to provide timely interventions. 
This paper evaluates the effectiveness of the deep learning transformer-based models BERT, RoBERTa, DistilBERT, ALBERT, and ELECTRA and various Long Short-Term Memory (LSTM) based models in detecting suicidal ideation from user posts on Reddit. 
%identify suicidal posts from Reddit utilizing transformer-based models. 
Toward this objective, we curated an extensive dataset from diverse subreddits and conducted linguistic, topic modeling, and statistical analyses to ensure data quality. Our results indicate that each model could reach high accuracy and F1 scores, but among them, RoBERTa emerged as the most effective model with an accuracy of 93.22\% and F1 score of 93.14\%. 
%In addition, we tested an ensemble method that combined model predictions through majority voting and turned out to be very competitive, though the overall achievements were slightly downward compared to RoBERTa. 
An LSTM model that uses attention and BERT embeddings performed as the second best, with an accuracy of 92.65\% and an F1 score of 92.69\%. 
Our findings show that transformer-based models have the potential to improve suicide ideation detection, thereby providing a path to develop robust mental health monitoring tools from social media. This research, therefore, underlines the undeniable prospect of advanced techniques in Natural Language Processing (NLP) while improving suicide prevention efforts.
\end{abstract}

%% file: Sections/introduction.tex
\section{Introduction}
Suicide is a leading global public health problem despite prevention efforts and strategies \cite{WorldHealth}. It is recognized not only as an individual phenomenon but also as one shaped by behavioral and environmental factors~\cite{gvion2012suicide}. This phenomenon is responsible for the annual death of more than 700,000 people around the globe, and many more individuals attempt suicide annually \cite{WorldHealth2}.
Moreover, it is the third preeminent reason for death among young American adults aged 20-24 years old \cite{VeryWellHealth}.

Recent studies have found that social media works as a tool for individuals undergoing suicide-related thoughts and behavior to share their psychological conditions where a plethora of posts uncover their suicidal thoughts, history, or intent ~\cite{gvion2012suicide}~\cite{de2016discovering}. Online media makes it easier for people to communicate, explain themselves, and create a record volume of user-generated content~\cite{sarker2015utilizing}. Communities like “SuicideWatch” on Reddit act as windows through which users express their most personal experiences, including psychological health issues. As a consequence, the development of suicidal ideation detection from such posts is important for timely intervention and support. More specifically, discovering suicidality patterns among people experiencing suicidal ideation is a noteworthy real-world problem to encounter which would eventually offer suicidal-prone folks a way of help and treatment. This motivated us to conduct this research exploring suicidal ideation in Reddit by utilizing state-of-the-art machine-learning approaches. Our goal is to evaluate the possibility of using deep learning transformers and LSTM models in detecting suicidal ideation posts.

The process of identifying posts with suicidal ideation from social media content suffers acutely from different challenges due to the nuanced and heterogeneous ways in which suicidal thoughts can be expressed. Most importantly, social media platforms such as Reddit and X/Twitter contain an enormous volume of information produced every second worldwide. In addition to this overloaded data, these contents can often be unstructured, noisy, and incomplete, posing a major challenge to data mining tools. To overcome these challenges, many traditional approaches to text classification based on association-based systems and classical machine learning algorithms have been experimented with~\cite{Shawkat}~\cite{hasan2021survey}. Such methods have enjoyed some success, however, most such attempts at text classification get stumped by the complexity and subtlety of natural language. Recent deep learning advances, especially transformer-based models like Bidirectional Encoder Representations from Transformers (BERT), have done at least a good job of understanding context and semantics in text~\cite{li2022survey}~\cite{gonzalez2020comparing}. These models can capture such fine-grained patterns, 
which potentially turn them into powerful tools for detecting suicidal ideation. Despite achieving better performance over traditional machine learning models, contemporary research shows that BERT-based approaches may have limitations in classifying specific kinds of documents~\cite{gao2021limitations}. This led us to carry out research that explores suicidal ideation in a large dataset collected from Reddit.
%by comparing transformer-based model variants on this dataset. 
%while eliminating various model selection criteria like performance metrics, document category, etc.
Our research study aims to prove the efficiency of transformer-based models in detecting posts with suicidal intent using a dataset of Reddit posts. The task is formulated to solve a binary classification problem. The purpose is to classify every post based on whether or not it contains suicidal intent. Since this is a very critical task, the precision and reliability of predictions need to be ensured. Furthermore, we examine diverse LSTM models with various word embeddings to provide a strong ground for BERT-based pre-trained models.

Our approach has two major objectives: first, the performance of deep learning transformer-based models in detecting suicidal intent posts on social media will be assessed, and second, the performance of LSTM-based models will be compared.
%firstly, assessment of the performance of models using transformers in detecting suicidal intent posted on social media, and secondly, to establish whether ensemble methods are better at combining strengths from several models. 
Designed methods for meeting these objectives contribute to developing practical tools for mental health monitoring on social media platforms and are a step forward toward suicide intervention.

\textbf{\textit{The primary contributions of the research presented in this paper are:}}

\begin{enumerate}
    \item Preparing a large annotated dataset from Reddit consisting of suicidal and non-suicidal posts. We make the dataset available to researchers upon request.
    \item Implementing statistical judgemental analysis and Latent Dirichlet Allocation (LDA) topic modeling on the suicidal posts to validate annotation accuracy.
    \item Evaluating the performance of multiple transformer-based classification models such as BERT and RoBERTa with the annotated dataset.
    \item Evaluating the efficacy of using LSTM-based models for suicidal ideation detection using a variety of text embedding techniques.
\end{enumerate}

%% file: Sections/related_work.tex
\section{Related work}
The identification of suicidal ideation (SI) from information shared on social media platforms has been a subject of attention in the past few years, driven by the increasing realization of early intervention possibilities for detecting mental health crises. This section overviews existing literature on traditional and deep learning approaches for automatically detecting suicidal intent from the text of social media posts. 

Early work in this area largely made use of classical machine learning techniques, such as Support Vector Machines (SVN), Naive Bayes, and Random Forests~\cite{haque2022comparative}~\cite{aldarwish2017predicting}. Such methods typically rely on the use of hand-engineered features, including linguistic cues, sentiment analysis, and lexical patterns indicative of suicidal ideation. De Choudhury et al.~\cite{de2013predicting} employed linguistic and behavioral features of Twitter to identify at-risk individuals for depression. They built an SVM classifier that predicted depression with 70\% accuracy. Tsugawa et al.~\cite{tsugawa2015recognizing} used a bag of words and other features to identify depression in Japanese text. They achieved 69\% accuracy using the SVM classifier.

Even though performance indicates the potential of classical machine learning methods for SI detection, these approaches depend heavily on extracted features, which cannot cover the diversity of all-natural language expressions of conditions related to the mental state~\cite{kowsari2019text}. With the rise of deep learning, various neural network architectures, such as Convolutional Neural Networks (CNNs) and Recurrent Neural Networks (RNNs), have been employed. These models learn representations automatically from raw data, removing the overhead of feature engineering. Sinha et al.~\cite{sinha2019suicidal} implemented deep learning sequential models and graph neural networks based on Twitter suicidal posts, user activities, and social networks formed among users posting about suicide in their study to explore suicidal ideation. Schoene et al.~\cite{schoene2021hierarchical} used hierarchical multi-scale RNNs in their research, introducing a dilated LSTM learning model to perform linguistic analysis in detecting suicidal notes collected from social media blogs. Similarly, Haque et al.~\cite{haque2022comparative} used a stack of CNNs and a Bidirectional long short-term memory (CNN–BiLSTM) model to identify local and sequential patterns in text when detecting suicidal ideation on Reddit data. Although improvements in the performance of these deep learning models over classical machine learning methods were achieved, the broader context and semantics of the text, which are crucial for the indication of subtle expressions of suicidal intent, remain lacking.

Models based on transformers, such as BERT by Devlin et al.~\cite{devlin2019bert}, have brought breakthroughs in deep learning by using self-attention mechanisms that acquire complex dependencies and context information in the text. This transformer-based model has been modified for almost all NLP tasks, as it can create corresponding contextual word embeddings. Qasim et al.~\cite{qasim2022fine} applied BERT-based models and their variants on three separate Twitter datasets to check the performance of transfer learning models, which showed prominent results. One recent research applied deep learning and attention-based models to relatively small datasets like \textit{UMD Reddit Suicidality Dataset}\footnote{\url{https://users.umiacs.umd.edu/~resnik/umd_reddit_suicidality_dataset.html}} where the accuracy ranged from 45\% to 60\%~\cite{long2022quantitative}. Moreover, Singh et al.~\cite{singh2023attention} used BERT-based embeddings with a BiLSTM network in detecting depression potential from social media, and their model was quite good at detecting symptoms of depression in cross-domain settings with accuracy close to 90\%.

Our work focuses on evaluating diverse transformer-based models to discriminate suicidal from non-suicidal posts on Reddit. We also investigate the benefits of LSTM-based models for this particular task. We make a few vital contributions towards this end compared to previous studies. First, we compile an extensive dataset from Reddit to train several BERT models and their variants to enable a comprehensive comparative analysis of their performance in detecting suicidal posts. Second, we implement several LSTM models with distinctive text embedding for suicidal detection systems. We compare the performance of these models against other BERT-based models. 

%% file: Sections/problem_definition.tex
\section{Problem Definition}
The primary objective of this study is to distinguish social media posts collected from Reddit that share suicidal intentions by employing transformer-based deep-learning models. We contemplate this problem of identifying suicidal ideation as a binary classification problem and define it as below:

\textbf{Problem} - \textit{
Given an annotated dataset D = \{$d_1$, $d_2$, ..., $d_n$\} of $n$ Reddit posts, the study aims to train text classification models capable of classifying each post $d_i$ to the label $p_i \in \{0, 1\}$, where $0$ indicates non-suicidal and $1$ indicates suicidal.} 
%The label $p_i$ can be encoded as either suicidal or non-suicidal.

As we march forward to solve this problem, we have formulated two research questions, each pointing towards the performance of classifying models. The questions are as follows:

\begin{enumerate}
    \item How competitive are transformer-based models in identifying suicidal posts?
    \item How is the performance of various LSTM-based models affected by integrating different embeddings (e.g. BERT, GloVe, and Word2Vec)?
\end{enumerate}

%% file: Sections/methodology.tex
\section{Methodology}
\begin{figure}
    \centering
    \includegraphics[width=1\linewidth]{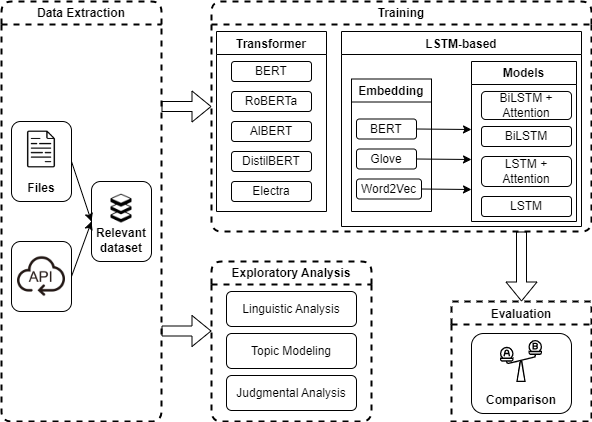}
    \caption{A summary of our research framework}
    \label{fig:research_architecture}
\end{figure}

%Our research aims to compare the performance of various transformer models in predicting suicidal content. 
Figure~\ref{fig:research_architecture} demonstrates a generic overview of our research architecture beginning from data collection and ending with model evaluation.

\input{Sections/Methodology/dataset}
\input{Sections/Methodology/data_processing}
\input{Sections/Methodology/exploratory_analysis}
\input{Sections/Methodology/classification_models}

%% file: Sections/Methodology/dataset.tex
\subsection{Dataset}

\subsubsection{Data Collection}
In our study, we employed Pushshift\footnote{https://pushshift.io/} API to collect available social media posts from Reddit over three months from October 2022 to December 2022. Our data collection is related to both suicidal and non-suicidal ideation ranging over diverse subreddit posts including \textit{r/SuicideWatch, r/socialanxiety, r/TrueOffMyChest, r/bipolar, r/confidence, and r/geopolitics.} We have selected these subreddits based on several aspects. Most importantly, we prioritized the total number of group members and their active membership which would allow us to get significant relevant data for model training. Moreover, we considered the subreddits providing enough text data to classify into two labels: suicidal and non-suicidal as well as make a balance between the two classes. We amassed 37821 posts, which were then annotated to the two labels.

\subsubsection{Annotation}
We have considered the posts from \textit{r/SuicideWatch} as suicidal since this subreddit is dedicated to users struggling with suicidal thoughts. Since this subreddit provides an open platform for discussing suicidal-prone circumstances (e.g. attempts, drug consumption, plans, and risky expressions), it is conceivable to expect that the authors have experienced suicidal intent in their lives. We have collected 18566 posts from this subreddit over these three months and annotated them as suicidal. In contrast, other subreddits have their distinct purposes like discussing politics, anxieties, and confessions, which are not related to suicidal factors at all. Thus, we have tagged these posts as non-suicidal. To validate our labeling, we performed both topic modeling and statistical judgmental analyses as discussed in sections IV.C.2 and IV.C.3.  

%% file: Sections/Methodology/data_processing.tex
\subsection{Data Processing}

This study followed an array of data processing steps in sequential order to make raw data feasible for data analysis. These steps primarily included white-space tokenization, cleaning, and normalization. Using regular expressions, the normalization stage removed unnecessary tokens including emojis, URLs, special characters, and digits. All of these stages were followed by the lemmatization process to group the inflected forms of a word and convert them to their root form, amplifying the representation of text attributes.

While conducting our research, we utilized data sets from separate data preparation stages. For instance, during Parts-of-Speech (POS) extraction, we used only tokenized data sets to keep the text as close to its original form as possible to preserve the context and structure the tagger relies upon whereas, during topic modeling, we utilized the final lemmatized data. Again, to extract suicidal phrases, we employed the original data set in the algorithm to receive meaningful phrasal words.

%% file: Sections/Methodology/exploratory_analysis.tex
\subsection{Exploratory Analysis}
We explore our collected Reddit dataset across two distinct phases. The first phase starts with textual analysis executed on both suicidal and non-suicidal data to discover valuable insights. The second phase involves topic modeling over the targeted subreddit to inspect the nature of discussions among users, determining how suicidal their shared contents are.

\subsubsection{Linguistic Analysis}
Here we are interested in identifying a bag of phrases that has a high probability of being available in suicidal communication. For suicidal phrase extraction, we took advantage of the TextRank~\cite{mihalcea-tarau-2004-textrank} algorithm implemented in the PyTextRank\footnote{https://pypi.org/project/pytextrank/} library. TextRank considers the relationships among words within a document and assigns a score to each term, highlighting the most informative terms that capture the essence of the document. We have extracted an array of distinguishable phrases that can be prevalent in suicide-related content. Some of these phrases are \textit{so much pain, goodbye, no hope, suicidal thought, mental illness, my suicide note, severe social anxiety, my first attempt, no other way, just a burden, self hatred, and my own death}.

\input{Tables/suicidal_data_stat}

Table~\ref{tab:suicidal_stat} exhibits an overview of the linguistic analysis of our data set. We can notice that although the average length of a suicidal post is 851.243 characters and token counts are 77.022, there is negligible usage of URLs and hashtags within these posts. This is also true for non-suicidal posts. Besides, the average non-suicidal posts tend to be longer than their counterparts. Other than this, we have observed the frequent use of curse words in those posts, for instance, f**k, ass**le, bull*hit, b*tch, and f**ker.

Furthermore, to calculate the POS tagging of our collected posts, we utilized the Penn Treebank~\cite{marcus1993building} implemented in the NLTK library. Table~\ref{tab:suicidal_stat} includes the average count of numerous POS tags.

\subsubsection{Topic Modeling}
We integrated topic-modeling techniques to examine the discussions among the \textit{r/SuicideWatch} community users i.e. how users vary their wording in sharing their thoughts and how suicidal their posts are. This helps to validate our labeling of these posts as suicidal. We applied Latent Dirichlet Allocation (LDA) topic modeling to study that relation. We selected LDA primarily because it is an unsupervised machine-learning model, therefore, we are not required to do any prior classification of the documents. Along with that, LDA has grown in popularity among the research community~\cite{jelodar2019latent}~\cite{islam2021mining} and showcased a significant resurgence among online communities involved with mental discomfort~\cite{Carron-Arthur2016}. 

We trained a topic model on the posts to observe the underlying shared topics within \textit{r/SuicideWatch} users. We combined the title and content for each post during this analysis. We utilized two prominent Python libraries, GENSIM\footnote{https://pypi.org/project/gensim/} and scikit-learn\footnote{https://scikit-learn.org/} for our LDA topic model implementation. Initially, we extracted the bigram and trigram phrases of the processed tokens using GENSIM. After obtaining bigram and trigram phrase tokens, we constructed a Term Frequency Inverse Document Frequency (TF-IDF) matrix. Finally, we fed these TF-IDF scores into the LDA model provided by scikit-learn with 1, 2, 3, ..., and 30 topics.

% We finalized the topic model used in our inspection based on their coherence scores, a measure of the semantic similarity between repetitive words in a topic. In our case, we received 

\begin{figure}
    \centering
    \includegraphics[width=0.8\linewidth]{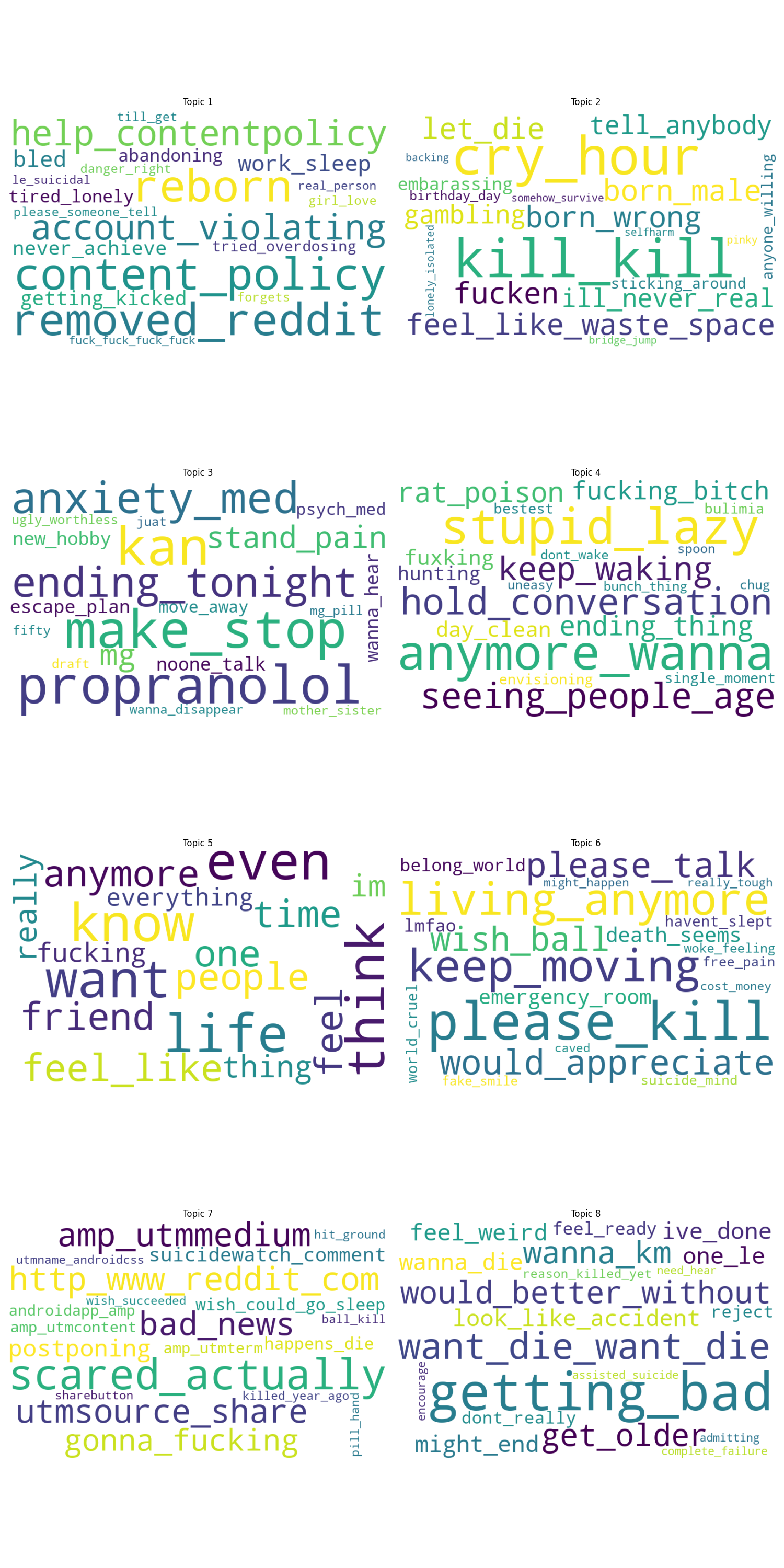}
    \caption{Results of Topic Modeling with 8 Topics}
    \label{fig:topic_modeling}
\end{figure}

Figure~\ref{fig:topic_modeling} demonstrates the result of our LDA topic model with a sample of eight topics. We can see each topic contains suicidal keywords like \textit{kill, ending\_tonight, make\_stop, and want\_die}. These terms suggest that the primary objective of the shared posts in the subreddit \textit{r/SuicideWatch} aligns with a similar intention. 
%This observation has directed us to consider the posts from this particular community as suicidal.

\subsubsection{Judgmental Analysis}
On top of topic modeling, we incorporated a human judgmental approach to our dataset to validate our annotation quality which is common practice among research communities~\cite{hasan2021survey}~\cite{chakraborty2024comparative}. To accomplish that, We randomly chose a sample of 756 posts, i.e. 2\%, from our large dataset. We manually annotated this sample dataset and evaluated the judgmental score (Cohens' Kappa, $\kappa$). The idea behind this approach is that if we find an accepted value of $\kappa$, we can say that this is approximately true for the whole dataset \cite{Landis1977}. By following this approach, our annotation policy is more plausible. 

For manual annotation, we defined a set of principles to assist in classifying the posts into two labels following the guidelines:

\begin{enumerate}
    \item \textbf{Posts with Suicidal Ideation}
        \begin{itemize}
            \item Post emphasizes a consequential outcome of suicidal intention. For instance, text fragments such as \textit{no longer have anything to hold on to} and \textit{I don't want to be known, and I just want to be gone}.
        
            \item Content includes former suicidal attempts or discussion of instant suicidal plans like \textit{If I can find a way, it ends tonight} and \textit{last night I attempted}.
 
        \end{itemize}
    \item \textbf{Posts without Suicidal Ideation}
        \begin{itemize}
            \item Post shares distressing or suicide-related news or information. For example, news about the suicide of a celebrity or a friend. 
            \item Post discusses any worldly matters except suicidal issues such as sports, politics, and traveling.
        \end{itemize}
\end{enumerate}

Both Cohen's kappa values of Author 1, 0.897, and Author 2, 0.854, agree within the near-perfect agreement, as conventionally, values above 0.8 are taken to express this concordance level. Indeed, this interpretation aligns with the guidelines of Landis et al.~\cite{Landis1977}, which set kappa values between 0.81 and 1.00 at almost perfect agreement.

%% file: Tables/suicidal_data_stat.tex
\begin{table}
    \caption{An overview of linguistic analysis of dataset}
    \centering
    \begin{tabular}{|c||cc|}
        \hline
         \textbf{Linguistic} & \multicolumn{2}{|c|}{\textbf{Values}} \\
        \cline{2-3}
         \textbf{Attributes} & \textbf{Suicidal Posts} & \textbf{Non-suicidal Posts} \\
        \hline
        Avg hashtags & 0.034 & 0.048 \\
        \hline
        Avg posts containing URL & 0.002 & 0.022 \\
        \hline
        Avg character length of posts & 851.243 & 1053.305 \\
        \hline
        Avg tokens & 77.022 & 94.040 \\
        \hline
        Avg nouns & 30.855 & 38.539 \\
        \hline
        Avg pronouns & 26.640 & 32.339 \\
        \hline
        Avg verbs & 41.169 & 46.872 \\
        \hline
        Avg Adjectives & 13.364 & 15.812 \\
        \hline
    \end{tabular}
    \label{tab:suicidal_stat}
\end{table}

%% file: Sections/Methodology/classification_models.tex
\subsection{Classification Models}
\label{subsec:classificaiton_models}

To gain literal knowledge in the unprocessed subreddit posts, we employed diverse state-of-the-art transformer-based text classification models and, utilized the models' outputs to get the result from our proposed ensemble model. Our trained classifiers are as follows:

\begin{itemize}
    
    \item \textbf{BERT:} BERT is a well-known model for any Natural Language Processing (NLP) task. It follows a bidirectional approach to learn contextual relations between words in a text to understand a word based on its surroundings from both directions~\cite{devlin2019bert}.
    
    \item \textbf{RoBERTa:} RoBERTa is a variant of BERT optimized with larger batches over more training data, longer sequences, and removing the next sentence prediction objective, which tuned key hyper-parameters to improve the pre-training procedure of BERT~\cite{Liu2019RoBERTaAR}.
    
    \item \textbf{DistilBERT:} The model DistilBERT is a distilled version of BERT, lighter and more speedy, whereby most of BERT's accuracy is maintained while becoming more computationally efficient~\cite{Sanh2019DistilBERTAD}. Additionally, it uses knowledge distillation, where knowledge supporting BERT is compressed into a smaller model.
    
    \item \textbf{ALBERT:} ALBERT is a lightweight version of BERT that reduces the size and complexity of the model, which in turn helps to make it light and quick without much loss of performance by its factorizing of the embedding parameters and sharing parameters across layers~\cite{Lan2019ALBERTAL}.
    
    \item \textbf{ELECTRA:} The model ELECTRA introduces a new pre-training approach to train the model for actual inputs against those replaced by a generator; this makes the method of sample efficiency stronger than BERT's Masked Language Modeling (MLM)~\cite{Clark2020ELECTRA}.
    
    \item \textbf{LSTM:} We implemented multiple LSTM models with and without attention levels to capture relevant context over a long period using a word's left and right contexts. Levels of attention enable the model to learn the pertinent words of a sentence. Again, we also tried our model training with and without bi-directional features. Each model has 100 LSTM units inside with a dropout rate of 0.25 and a recurrent dropout rate of 0.2. Another dropout layer of 0.2 was added after the attention layer upon existence. Two more fully connected layers were employed at the back end of the models having 256 and 2 units, respectively. We used several context-aware pre-trained embedding provided by BERT, Glove, and Word2Vec for our word embedding layer.
\end{itemize}

% \subsection{Proposed Ensemble Model}
% \label{subsec:proposed_ensemble_model}

% In this study, we propose a voting ensemble model that combines the predictions of all transformer-based models for a given text and returns a label with the highest appearances. We hypothesize that an ensemble scheme on model combinations would provide reliable performance in predicting text labels. This would highlight the best-performed model outputs and simultaneously eliminate the number of model selection criteria. Our proposed ensemble scheme is formulated as below:
% %Remark: why would we worry about a number of model selection criteria? Our experimental results show that RoBERTa performed best, therefore, that is the model that should be used. Therefore, I believe that the last sentence in the above paragraph is irrelevant and should be removed.

% \textbf{Ensemble Scheme} - \textit{
% Given an array of predicted results P~=~\{$p_1$, $p_2$, ..., $p_m$\} of a data instance $d_i$ where $m$ denotes the model count, the ensemble scheme results in a single predicted label $l_i$ that occurs most frequently among the predictions of all models. The finalized label $l_i$ can be written as:
% }

% \begin{equation}
%     l_i = \operatorname{argmax}_p \sum_{k=1}^{m} F(p_k = p)
% \end{equation}

% \textit{
% where:
% \begin{itemize}
%     \item $p \in \{suicidal, non-suicidal\}$
%     \item $p_k$ denotes the predicted label by the model $k \in \{1,2, ..., m\}$
%     \item F symbolizes the indicator function that equals 1 if the argument is true and 0 otherwise
% \end{itemize}
% }

%% file: Sections/results.tex
\section{Experiments and Results}
To ensure that each fold has a balanced class distribution for all our experiments, we applied a 5-fold stratified cross-validation. Initially, we divided the whole dataset into 20\% to test the models' overall performance and the remaining 80\% to use as a train-validation split during the cross-validation training process. All the models were trained and evaluated on each of the 5 train-validation splits and hence, we trained 5 versions of each classification model, one for each respective fold. Furthermore, we tuned each model using a grid search named Ray-Tune %\footnote{https://www.ray.io/} 
over the validation datasets~\cite{Ray}. We found the best-performed models for the learning rate $10^{-6}$ and weight decay $10^{-2}$ in the case of every model. During the training of our deep learning text classifiers, the AdamW optimizer~\cite{loshchilov2018decoupled} was used along with a learning rate scheduler having the patience until three epochs. We implemented these models using the functionalities provided by PyTorch
and HuggingFace. The models were trained on a system with two NVIDIA Ampere A100 GPUs (6912 CUDA cores, 40GB RAM each) and 512 GB RAM.

To predict the class of a test sample using a specific model, we combine the predicted classes from the model's 5 trained folds generated through the 5-fold cross-validation and return the class with the highest recurrence. These outputs are then utilized to evaluate model performance. In Table~\ref{tab:model_performance}, we report the scores of the performance metrics (i.e. Accuracy, F1 score, Precision, and Recall) for each of our experimented models.

\input{Tables/model_performance}

\subsection{How competitive are transformer-based models in identifying suicidal ideation posts?}

To answer this question, we must analyze the performance of the transformer-based deep learning models evaluated in our experiments. The results shown in Table~\ref{tab:model_performance} indicate the following:
\begin{itemize}
    \item All models achieved impressive results on our large dataset with average cross-validation accuracy and F1 score above 91\%. This shows the effectiveness of transformer-based models in identifying suicidal posts on social media.
    
    \item RoBERTa performed best with 93.22\% average cross-validation accuracy and 93.14\% F1 score.
    
    \item ELECTRA and BERT achieved results above 92\%. These results are competitive with the best model, RoBERTa, and fall within a 1\% difference from RoBERTa. 
    \item All models gave excellent F1-score values above 91\% with a balanced performance concerning precision and recall. This indicates that the models correctly classify both negative and positive instances of suicidal posts, hence reducing false positives and false negatives. 
    \item Although not as high as the other models, the results from DistilBERT and ALBERT are still very respectable, at accuracies of 91.95\% and 92.17\%, respectively. 
    \item One noticeable performance by DistilBERT is that it took almost half the time to train compared with other variants, and still produced competitive results. So, if training time is critical and one is presented with a large dataset, DistilBERT is a good choice.
\end{itemize}

\subsection{How is the performance of various LSTM-based models affected by integrating different embeddings such as BERT, GloVe, and Word2Vec?}

The kind of text embeddings used with LSTM-based models highly influenced their performance. Table~\ref{tab:model_performance} shows that all LSTM-based models with BERT embeddings performed better than GloVe and Word2Vec-based models in each evaluation metric. Specifically, 
%BERT LSTM models work even better than others; namely, 
the BERT LSTM + Attention model attains an accuracy of 92.65\%, an F1 score of 92.69\%, a precision of 90.61\%, and a recall of 94.86\%. Other combinations achieved similar results.
%at least as good: BERT BiLSTM and BERT BiLSTM + Attention models show very high metrics and clearly confirm the high robustness of BERT text embeddings in improving model performance.

In contrast, models that use GloVe embeddings have a steep decline of about 14\% to 20\% concerning each performance metric. For example, the Glove BiLSTM + Attention model has an accuracy of 76.55\%, F1 score of 75.20\%, precision of 78.21\%, and recall of 72.40\%. This is a consistent trend across all GloVe LSTM variants.
%, no doubt further illustrating that whereas, GloVe embeddings do afford a fair improvement, their performance level is still way below that of BERT embeddings. 
LSTM-based models that used Word2Vec embeddings performed the worst, further showing the superiority of BERT embeddings over GolVe and Word2Vec.  
%For example, the Word2Vec LSTM + Attention model has an accuracy of only 53.67\%, F1 score of 54.64\%, precision of 55.42\%, and recall of 53.87\%. Even with BiLSTM or attention mechanism, performance is far worse than with BERT or GloVe embeddings.

BERT text embedding appears to be superior to other word embeddings (i.e., Glove and Word2Vec) because 1) BERT is pre-trained bidirectionally whereas, others follow skip-grams and a continuous bag of words (CBOW), 2) BERT considers the entire sentence to determine the embedding of each word whereas in others each word has a single representation regardless of its context in the sentence, and 3) BERT is designed for fine-tuning so that it can be adapted for a particular task. In contrast, others have limitations in adapting to specific tasks~\cite{Wang2021} \cite{ekmanBook}.
Our results support other works that show that BERT embeddings outperform other word embedding techniques for various NLP tasks~\cite{Toshevska}.

%The findings of this paper are such that overall, after integrating various embeddings into LSTM-based models in NLP tasks, the use of BERT embeddings boosts the models to a great extent, followed by GloVe, and then Word2Vec embeddings. Therefore, this shows that for better results in NLP models, one should use embeddings such as BERT embeddings to achieve the best performance.

%% file: Tables/model_performance.tex
\begin{table*}[htb]
    \caption{
    Summary of model performance with the evaluation metrics Accuracy, F1 score, Precision, and Recall \\
    (Training time may vary depending on the machine)
    }
    \centering
    \begin{tabular}{cc|cccc|c}
        \hline
        \textbf{Embedding} & \textbf{Model} & \textbf{Accuracy} & \textbf{F1} & \textbf{Precision} & \textbf{Recall} & \textbf{Approx. Time for each best model (hours)} \\
        \hline
        \hline
        
        RoBERTa & RoBERTa & \textbf{93.219} & \textbf{93.143} & \textbf{92.882} & 93.405 & 1.504 \\
        \hline
        
        DistilBERT & DistilBERT & 91.950 & 91.866 & 91.536 & 92.198 & \textbf{0.75} \\
        \hline
        
        ALBERT & ALBERT & 92.174 & 92.083 & 91.862 & 92.306 & 1.62 \\
        \hline
        
        BERT & BERT & 92.386 & 92.304 & 92.009 & 92.601 & 1.49 \\
        \hline
        
        ELECTRA & ELECTRA & 92.571 & 92.542 & 91.618 & 93.485 & 1.50 \\
        \hline
        \hline

        BERT & BiLSTM + Attention & 92.307 & 92.352 & 90.195 & 94.615 & 1.53 \\
        \hline

        BERT & BiLSTM & 92.531 & 92.543 & 90.758 & 94.40 & 1.52 \\
        \hline

        BERT & LSTM + Attention & 92.650 & 92.686 & 90.612 & \textbf{94.857} & 1.52 \\
        \hline

        BERT & LSTM & 92.690 & 92.682 & 91.127 & 94.292 & 1.51 \\
        \hline
        \hline

        GloVe & BiLSTM + Attention & 76.550 & 75.196 & 78.214 & 72.402 & 8.13 \\
        \hline

        GloVe & BiLSTM & 76.356 & 75.126 & 78.115 & 72.357 & 7.98 \\
        \hline

        GloVe & LSTM + Attention & 66.821 & 65.255 & 67.151 & 63.463 & 8.23 \\
        \hline

        GloVe & LSTM & 66.637 & 65.014 & 66.987 & 63.153 & 8.16 \\
        \hline
        \hline

        Word2Vec & BiLSTM + Attention & 64.759 & 65.112 & 66.835 & 63.476 & 7.31 \\
        \hline

        Word2Vec & BiLSTM & 64.837 & 64.834 & 66.137 & 63.582 & 7.19 \\
        \hline

        Word2Vec & LSTM + Attention & 53.673 & 54.635 & 55.419 & 53.873 & 7.06 \\
        \hline

        Word2Vec & LSTM & 53.815 & 54.694 & 55.735 & 53.691 & 6.93 \\
        \hline
        \hline
        
        % & Ensemble & 93.113 & 93.047 & 92.639 & 93.458 & -- \\
        \hline
    \end{tabular}
    \label{tab:model_performance}
\end{table*}

%% file: Sections/conclusion.tex
\section{Conclusion and future work}

In this paper, we evaluated the performance of five deep-learning transformer models along with LSTM variants in detecting suicidal ideation in a large dataset that we collected from Reddit. Based on our work, we make the following concluding remarks: 
\begin{enumerate}
    \item Though RoBERTa performed the best, all models gave competitive results within 1.5\% of RoBERTa in terms of each performance metric. Furthermore, the accuracy and F1 score results of all models were above 91\%, which showcases their competitiveness in detecting posts with suicidal ideation. 
    \item The light model DistilBERT was trained in about half the time to train other models, which makes it a good option to train on large datasets.  
    \item LSTM models using BERT embeddings achieved excellent results competitive with transformer models evaluated in our research. Moreover, BERT embeddings resulted in better LSTM models than GloVe and Word2Vec embeddings to detect posts with suicidal intent. 
\end{enumerate}

For future work, we plan to introduce an ensemble approach for suicidal ideation detection. We also plan to acquire a wider range of suicide datasets and evaluate our models on these datasets. This can include multilingual datasets with diverse cultural backgrounds to make the models effective globally. Furthermore, we want to research user-focused learning that explores user history and behavioral patterns to enhance suicidal detection accuracy.

%% file: main.bbl
\begin{thebibliography}{10}
\providecommand{\url}[1]{#1}
\csname url@samestyle\endcsname
\providecommand{\newblock}{\relax}
\providecommand{\bibinfo}[2]{#2}
\providecommand{\BIBentrySTDinterwordspacing}{\spaceskip=0pt\relax}
\providecommand{\BIBentryALTinterwordstretchfactor}{4}
\providecommand{\BIBentryALTinterwordspacing}{\spaceskip=\fontdimen2\font plus
\BIBentryALTinterwordstretchfactor\fontdimen3\font minus \fontdimen4\font\relax}
\providecommand{\BIBforeignlanguage}[2]{{%
\expandafter\ifx\csname l@#1\endcsname\relax
\typeout{** WARNING: IEEEtran.bst: No hyphenation pattern has been}%
\typeout{** loaded for the language `#1'. Using the pattern for}%
\typeout{** the default language instead.}%
\else
\language=\csname l@#1\endcsname
\fi
#2}}
\providecommand{\BIBdecl}{\relax}
\BIBdecl

\bibitem{WorldHealth}
\BIBentryALTinterwordspacing
W.~H. Organization, \emph{National suicide prevention strategies: progress, examples, and indicators}, 2018. [Online]. Available: \url{https://www.who.int/publications/i/item/national-suicide-prevention-strategies-progress-examples-and-indicators}
\BIBentrySTDinterwordspacing

\bibitem{gvion2012suicide}
Y.~Gvion and A.~Apter, ``Suicide and suicidal behavior,'' \emph{Public health reviews}, vol.~34, pp. 1--20, 2012.

\bibitem{WorldHealth2}
\BIBentryALTinterwordspacing
W.~H. Organization, \emph{Suicide}, 2018. [Online]. Available: \url{https://www.who.int/news-room/fact-sheets/detail/suicide}
\BIBentrySTDinterwordspacing

\bibitem{VeryWellHealth}
\BIBentryALTinterwordspacing
M.~Stibich, \emph{Top 10 Causes of Death for Americans Ages 20 to 24}, 2024. [Online]. Available: \url{https://www.verywellhealth.com/top-causes-of-death-for-ages-15-24-2223960}
\BIBentrySTDinterwordspacing

\bibitem{de2016discovering}
M.~De~Choudhury, E.~Kiciman, M.~Dredze, G.~Coppersmith, and M.~Kumar, ``Discovering shifts to suicidal ideation from mental health content in social media,'' in \emph{Proceedings of the 2016 CHI conference on human factors in computing systems}, 2016, pp. 2098--2110.

\bibitem{sarker2015utilizing}
\BIBentryALTinterwordspacing
A.~Sarker, R.~Ginn, A.~Nikfarjam, K.~O’Connor, K.~Smith, S.~Jayaraman, T.~Upadhaya, and G.~Gonzalez, ``Utilizing social media data for pharmacovigilance: A review,'' \emph{Journal of Biomedical Informatics}, vol.~54, pp. 202--212, 2015. [Online]. Available: \url{https://www.sciencedirect.com/science/article/pii/S1532046415000362}
\BIBentrySTDinterwordspacing

\bibitem{Shawkat}
N.~Shawkat, J.~Saquer, and H.~Shatnawi, ``Evaluation of different machine learning and deep learning techniques for hate speech detection,'' in \emph{2024 ACM Southeast Conference (ACMSE 2024)}, Marietta, GA, USA, 2024, pp. 253--258.

\bibitem{hasan2021survey}
\BIBentryALTinterwordspacing
K.~Hasan, P.~Chakraborty, R.~Shahriyar, A.~Iqbal, and G.~Uddin, ``A survey-based qualitative study to characterize expectations of software developers from five stakeholders,'' in \emph{Proceedings of the 15th ACM / IEEE International Symposium on Empirical Software Engineering and Measurement (ESEM)}, ser. ESEM '21.\hskip 1em plus 0.5em minus 0.4em\relax New York, NY, USA: Association for Computing Machinery, 2021. [Online]. Available: \url{https://doi.org/10.1145/3475716.3475787}
\BIBentrySTDinterwordspacing

\bibitem{li2022survey}
Q.~Li, H.~Peng, J.~Li, C.~Xia, R.~Yang, L.~Sun, P.~S. Yu, and L.~He, ``A survey on text classification: From traditional to deep learning,'' \emph{ACM Transactions on Intelligent Systems and Technology (TIST)}, vol.~13, no.~2, pp. 1--41, 2022.

\bibitem{gonzalez2020comparing}
S.~González-Carvajal and E.~Garrido-Merchán, ``Comparing bert against traditional machine learning text classification,'' \emph{Journal of Computational and Cognitive Engineering}, vol.~2, 05 2020.

\bibitem{gao2021limitations}
S.~Gao, M.~Alawad, M.~T. Young, J.~Gounley, N.~Schaefferkoetter, H.~J. Yoon, X.-C. Wu, E.~B. Durbin, J.~Doherty, A.~Stroup \emph{et~al.}, ``Limitations of transformers on clinical text classification,'' \emph{IEEE Journal of Biomedical and Health Informatics}, vol.~25, no.~9, pp. 3596--3607, 2021.

\bibitem{haque2022comparative}
\BIBentryALTinterwordspacing
R.~Haque, N.~Islam, M.~Islam, and M.~M. Ahsan, ``A comparative analysis on suicidal ideation detection using nlp, machine, and deep learning,'' \emph{Technologies}, vol.~10, no.~3, 2022. [Online]. Available: \url{https://www.mdpi.com/2227-7080/10/3/57}
\BIBentrySTDinterwordspacing

\bibitem{aldarwish2017predicting}
M.~M. Aldarwish and H.~F. Ahmad, ``Predicting depression levels using social media posts,'' in \emph{2017 IEEE 13th International Symposium on Autonomous Decentralized System (ISADS)}.\hskip 1em plus 0.5em minus 0.4em\relax IEEE, 2017, pp. 277--280.

\bibitem{de2013predicting}
M.~De~Choudhury, M.~Gamon, S.~Counts, and E.~Horvitz, ``Predicting depression via social media,'' in \emph{Proceedings of the international AAAI conference on web and social media}, vol.~7, no.~1, 2013, pp. 128--137.

\bibitem{tsugawa2015recognizing}
S.~Tsugawa, Y.~Kikuchi, F.~Kishino, K.~Nakajima, Y.~Itoh, and H.~Ohsaki, ``Recognizing depression from twitter activity,'' in \emph{Proceedings of the 33rd annual ACM conference on human factors in computing systems}, 2015, pp. 3187--3196.

\bibitem{kowsari2019text}
K.~Kowsari, K.~Jafari~Meimandi, M.~Heidarysafa, S.~Mendu, L.~Barnes, and D.~Brown, ``Text classification algorithms: A survey,'' \emph{Information}, vol.~10, no.~4, p. 150, 2019.

\bibitem{sinha2019suicidal}
P.~P. Sinha, R.~Mishra, R.~Sawhney, D.~Mahata, R.~R. Shah, and H.~Liu, ``\# suicidal-a multipronged approach to identify and explore suicidal ideation in twitter,'' in \emph{Proceedings of the 28th ACM International Conference on Information and Knowledge Management}, 2019, pp. 941--950.

\bibitem{schoene2021hierarchical}
A.~M. Schoene, A.~Turner, G.~R. De~Mel, and N.~Dethlefs, ``Hierarchical multiscale recurrent neural networks for detecting suicide notes,'' \emph{IEEE Transactions on Affective Computing}, 2021.

\bibitem{devlin2019bert}
\BIBentryALTinterwordspacing
J.~Devlin, M.-W. Chang, K.~Lee, and K.~Toutanova, ``{BERT}: Pre-training of deep bidirectional transformers for language understanding,'' in \emph{Proceedings of the 2019 Conference of the North {A}merican Chapter of the Association for Computational Linguistics: Human Language Technologies, Volume 1 (Long and Short Papers)}.\hskip 1em plus 0.5em minus 0.4em\relax Minneapolis, Minnesota: Association for Computational Linguistics, Jun. 2019, pp. 4171--4186. [Online]. Available: \url{https://aclanthology.org/N19-1423}
\BIBentrySTDinterwordspacing

\bibitem{qasim2022fine}
R.~Qasim, W.~H. Bangyal, M.~A. Alqarni, and A.~Ali~Almazroi, ``A fine-tuned bert-based transfer learning approach for text classification,'' \emph{Journal of healthcare engineering}, vol. 2022, no.~1, p. 3498123, 2022.

\bibitem{long2022quantitative}
S.~Long, R.~Cabral, J.~Poon, and S.~C. Han, ``A quantitative and qualitative analysis of suicide ideation detection using deep learning,'' in \emph{Proceedings of the HealTAC 2022: the 5th Healthcare Text Analytics Conference}, 2022.

\bibitem{singh2023attention}
J.~Singh, N.~Singh, M.~M. Fouda, L.~Saba, and J.~S. Suri, ``Attention-enabled ensemble deep learning models and their validation for depression detection: a domain adoption paradigm,'' \emph{Diagnostics}, vol.~13, no.~12, p. 2092, 2023.

\bibitem{mihalcea-tarau-2004-textrank}
\BIBentryALTinterwordspacing
R.~Mihalcea and P.~Tarau, ``{T}ext{R}ank: Bringing order into text,'' in \emph{Proceedings of the 2004 Conference on Empirical Methods in Natural Language Processing}, D.~Lin and D.~Wu, Eds.\hskip 1em plus 0.5em minus 0.4em\relax Barcelona, Spain: Association for Computational Linguistics, Jul. 2004, pp. 404--411. [Online]. Available: \url{https://aclanthology.org/W04-3252}
\BIBentrySTDinterwordspacing

\bibitem{marcus1993building}
M.~Marcus, B.~Santorini, and M.~A. Marcinkiewicz, ``Building a large annotated corpus of english: The penn treebank,'' \emph{Computational linguistics}, vol.~19, no.~2, pp. 313--330, 1993.

\bibitem{jelodar2019latent}
H.~Jelodar, Y.~Wang, C.~Yuan, X.~Feng, X.~Jiang, Y.~Li, and L.~Zhao, ``Latent dirichlet allocation (lda) and topic modeling: models, applications, a survey,'' \emph{Multimedia tools and applications}, vol.~78, pp. 15\,169--15\,211, 2019.

\bibitem{islam2021mining}
S.~Islam, K.~Hasan, and R.~Shahriyar, ``Mining developer questions about major nosql databases,'' \emph{Int. J. Comput. Appl}, vol. 975, p. 8887, 2021.

\bibitem{Carron-Arthur2016}
\BIBentryALTinterwordspacing
B.~Carron-Arthur, J.~Reynolds, K.~Bennett, A.~Bennett, and K.~M. Griffiths, ``What’s all the talk about? topic modeling in a mental health internet support group,'' \emph{BMC Psychiatry}, vol.~16, no.~1, October 2016. [Online]. Available: \url{https://doi.org/10.1186/s12888-016-1073-5}
\BIBentrySTDinterwordspacing

\bibitem{chakraborty2024comparative}
P.~Chakraborty, K.~Hasan, A.~Iqbal, G.~Uddin, and R.~Shahriyar, ``A comparative study of software development practices in bangladesh, an emerging country,'' \emph{International Journal of Software Engineering, Technology and Applications}, vol.~2, no.~2, pp. 149--187, 2024.

\bibitem{Landis1977}
\BIBentryALTinterwordspacing
J.~R. Landis and G.~G. Koch, ``The measurement of observer agreement for categorical data,'' \emph{Biometrics}, vol.~33, no.~1, pp. 159--174, 1977. [Online]. Available: \url{http://www.jstor.org/stable/2529310}
\BIBentrySTDinterwordspacing

\bibitem{Liu2019RoBERTaAR}
\BIBentryALTinterwordspacing
\emph{RoBERTa: A Robustly Optimized BERT Pretraining Approach}, vol. abs/1907.11692, 2019. [Online]. Available: \url{https://api.semanticscholar.org/CorpusID:198953378}
\BIBentrySTDinterwordspacing

\bibitem{Sanh2019DistilBERTAD}
\BIBentryALTinterwordspacing
V.~Sanh, L.~Debut, J.~Chaumond, and T.~Wolf, ``Distilbert, a distilled version of bert: smaller, faster, cheaper and lighter,'' 2020. [Online]. Available: \url{https://arxiv.org/abs/1910.01108}
\BIBentrySTDinterwordspacing

\bibitem{Lan2019ALBERTAL}
\BIBentryALTinterwordspacing
Z.~Lan, M.~Chen, S.~Goodman, K.~Gimpel, P.~Sharma, and R.~Soricut, ``Albert: A lite bert for self-supervised learning of language representations,'' 2020. [Online]. Available: \url{https://arxiv.org/abs/1909.11942}
\BIBentrySTDinterwordspacing

\bibitem{Clark2020ELECTRA}
\BIBentryALTinterwordspacing
K.~Clark, M.-T. Luong, Q.~V. Le, and C.~D. Manning, ``Electra: Pre-training text encoders as discriminators rather than generators,'' in \emph{International Conference on Learning Representations}, 2020. [Online]. Available: \url{https://openreview.net/forum?id=r1xMH1BtvB}
\BIBentrySTDinterwordspacing

\bibitem{Ray}
\BIBentryALTinterwordspacing
Ray, \emph{Effortlessly scale your most complex workloads}, 2024. [Online]. Available: \url{https://www.ray.io/}
\BIBentrySTDinterwordspacing

\bibitem{loshchilov2018decoupled}
\BIBentryALTinterwordspacing
I.~Loshchilov and F.~Hutter, ``Decoupled weight decay regularization,'' in \emph{International Conference on Learning Representations}, 2019. [Online]. Available: \url{https://arxiv.org/abs/1711.05101}
\BIBentrySTDinterwordspacing

\bibitem{Wang2021}
Y.~Wang, L.~Cui, and Y.~Zhang, ``Improving skip-gram embeddings using bert,'' \emph{IEEE/ACM Transactions on Audio, Speech, and Language Processing}, vol.~29, pp. 1318--1328, 2021.

\bibitem{ekmanBook}
M.~Ekma, \emph{Learning Deep Learning: Theory and Practice of Neural Networks, Computer Vision, Natural Language Processing, and Transformers Using TensorFlow}.\hskip 1em plus 0.5em minus 0.4em\relax Boston, MA, USA: Addison-Wesley Professional, 2021.

\bibitem{Toshevska}
M.~Toshevska, F.~Stojanovska, and J.~Kalajdjieski, ``Comparative analysis of word embeddings for capturing word similarities,'' \emph{arXiv preprint arXiv:2005.03812}, 2020.

\end{thebibliography}
